%% file: main.tex
\documentclass{article}

\PassOptionsToPackage{numbers,sort}{natbib}
\usepackage[dblblindworkshop,preprint]{neurips_2026}
\workshoptitle{Physical World AI}
\input{preamble}

\title{New Evidence, Same Choice:\\
Testing Physical Experiment Selection\\
in Vision Language Models}
\author{
Sourajit Saha$^{1}$ \quad Shubhashis Roy Dipta$^{1}$ \quad Nobin Sarwar$^{1}$ \quad Shaswati Saha$^{1}$\\[2pt]
{\bf Yuxuan Jiang$^{1}$ \quad Siyuan Li$^{2}$ \quad Qiheng Wang$^{3}$}\\[3pt]
$^{1}$University of Maryland, Baltimore County\\
$^{2}$University of Georgia\\
$^{3}$Independent Researcher\\
{\tt\small \{ssaha2, sroydip1, sms2, ssaha3, yuxuanj1\}@umbc.edu}\\
{\textcolor{blue}{\normalsize\url{https://sourajitcs.github.io/physicalprobe/}}}\\[2pt]
}

\begin{document}
\maketitle

\begin{center}
    \includegraphics[width=\linewidth]{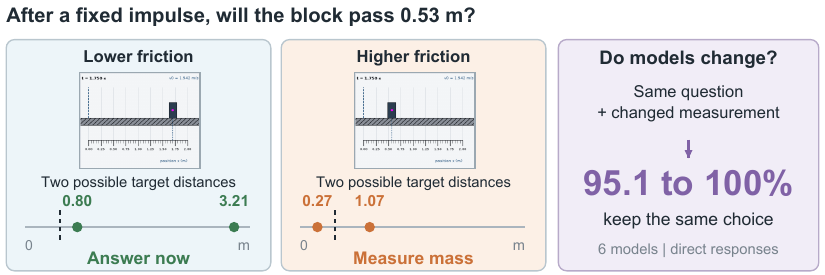}
    \captionof{figure}{
    \textbf{The same question can require a different experiment.}
    Two earlier coasting tests reveal different friction.
    The question asks about a new trial after a fixed impulse.
    Left: either mass passes the mark, so the answer is known.
    Middle: one possible distance falls short, so mass must be measured.
    Right: direct model choices usually fail to change across such pairs.
    This recorded example was selected for illustration, with its question shortened.
    }
    \label{fig:teaser}
\end{center}

\input{sec/01_abstract}
\input{sec/02_introduction}
\input{sec/03_related_work}
\input{sec/04_method}
\input{sec/05_experiments}
\input{sec/06_results}
\input{sec/07_conclusion}
\input{sec/08_limitations}

\label{page:mainend}

\begingroup
\raggedright
\renewcommand{\url}[1]{\href{#1}{source}}
\bibliographystyle{paperrefs}
\bibliography{literature, extra}
\endgroup

\clearpage
\appendix
\input{sec/09_appendix}
\FloatBarrier
\clearpage

\input{sec/10_prompt_example}
\end{document}

%% file: preamble.tex
\usepackage[T1]{fontenc}
\usepackage[utf8]{inputenc}
\usepackage{microtype}
\usepackage{graphicx}
\usepackage{booktabs}
\usepackage{multirow}
\usepackage{makecell}
\usepackage{amsmath,amssymb,mathtools}
\usepackage[table]{xcolor}
\usepackage{xspace}
\usepackage{enumitem}
\usepackage{caption}
\usepackage{float}
\usepackage{placeins}
\usepackage{needspace}
\usepackage{url}
\usepackage[hidelinks]{hyperref}
\usepackage{cleveref}

\definecolor{methodcolor}{HTML}{006B9E}
\definecolor{softblue}{HTML}{EEF5FA}
\definecolor{softorange}{HTML}{FFF0E6}
\definecolor{softgray}{HTML}{F2F2F2}
\newcommand{\stopaction}{\textsc{stop}\xspace}
\newcommand{\yes}{\textsc{yes}\xspace}
\newcommand{\no}{\textsc{no}\xspace}
\newcommand{\unknown}{\textsc{undetermined}\xspace}
\newcommand{\worlds}{\mathcal{W}}
\newcommand{\actions}{\mathcal{A}}

\DeclareMathOperator*{\argmin}{arg\,min}

\renewcommand{\paragraph}[1]{\smallskip\noindent\textbf{#1.}}
\setlist[itemize]{leftmargin=1.35em,itemsep=1pt,topsep=2pt,parsep=0pt}
\hypersetup{pdftitle={New Evidence, Same Choice},pdfauthor={},pdfsubject={Physical experiment selection},pdfkeywords={physical reasoning, vision language models, experiment selection}}

%% file: sec/01_abstract.tex
\begin{abstract}

Consider a simple mechanics task involving a sliding block, a bouncing object, or a mass attached to a spring.
A model first sees an image from one measurement experiment; for example, how far a block coasted and must answer a question about a new trial, such as whether the block will pass a target after a fixed push.
The first experiment may provide enough information to answer, or the model may need another measurement, such as the object's mass, friction, restitution, or spring stiffness.
We study whether vision language models can decide when to answer immediately and, when more evidence is needed, which experiment to perform.
Current physical reasoning benchmarks usually score only the final answer, so they do not directly evaluate this decision.
We introduce a controlled evaluation in which each problem presents one measurement image and four possible physical worlds formed by two possible masses and two possible values of another relevant property.
The model must either stop and answer or select the cheapest additional experiment that can resolve the question.
We construct matched problem pairs in which changing either the observed measurement or the question changes the correct action.
Because all possible worlds and experiment costs are known, we can determine the optimal action explicitly.
Across six open models and 144 physical parameter sets, direct responses repeat the same action for 95.1\% to 100\% of image pairs even though the correct action changes.
Brief reasoning leads to more action changes, but the best model makes both decisions correctly for only 5.9\% of image pairs.
Additional tests reveal errors in reading measurements, performing physical calculations, and following the required response format.
By evaluating evidence selection separately from final answers, our benchmark exposes failures that conventional answer accuracy can hide.

\end{abstract}

%% file: sec/02_introduction.tex
\section{Introduction}
\label{sec:intro}
The goal of physical experiment selection is to obtain the information needed for a particular prediction.
Consider a block whose surface friction is known.
Its mass affects how far it moves after a short push with a fixed impulse.
If every possible mass carries it past a requested mark, measuring mass serves no purpose.
For a farther mark, that measurement may become necessary.
The value of a test depends on the observation and the question together.

Image language representations and instruction tuning support a broad set of visual tasks~\citep{radford2021clip,li2023blip2,liu2023llava}.
Physical benchmarks examine future motion, hidden properties, and object interactions~\citep{bear2021physion,tung2023physionpp,yi2020clevrer,chow2025physbench}.
Interactive agents extend this setting by collecting their own observations~\citep{xu2023ivre,xu2025deepphy,chandra2026hexa,lin2026physcap}.
However, an average score can conceal a fixed decision rule~\citep{lia-etal-2025-read}.
Always stopping earns half the available credit when half the questions are already answerable.
Such a score gives no direct test of whether the agent responds to the evidence.

We address this gap through matched physical problems.
Two versions share the question, menu, and prices, but show different measurement results.
We choose the question so that one version needs an additional test and the other does not.
A second comparison holds the image fixed and changes the requested threshold.
Scoring both members together exposes failures that an isolated correct choice can hide.

Our evaluation spans sliding, bouncing, and spring motion.
The resulting behavior is not a single error pattern.
Some models request a property that was already measured.
Others stop while multiple answers remain possible.
Even a useful selected observation can be followed by an incorrect answer.
Models also score poorly on reading and calculation, so selection alone cannot explain these failures.

Our key contributions and findings include:
\begin{itemize}
    \item A paired evaluation whose finite physical possibilities determine when an experiment is needed and which test has minimum cost;
    \item Evidence that six open models rarely handle both sides of the constructed comparisons correctly; and
    \item Dedicated studies of measurement choice, answer use, visual reading, calculation, prompt format, and cost, with recorded examples across all three tasks.
\end{itemize}

%% file: sec/03_related_work.tex
\section{Related Work}
\label{sec:related}
\paragraph{Predicting physical events}
Intuitive physics models learn regularities of motion~\citep{battaglia2013simulation,piloto2022intuitive}.
Scene representations support inference of object properties~\citep{wu2015galileo,wu2017deanimation}, while controlled benchmarks test interactions and future outcomes~\citep{bakhtin2019phyre,yi2020clevrer,bear2021physion,riochet2018intphys}.
PhysBench broadens this evaluation to vision language models~\citep{chow2025physbench}.
Physion++ requires hidden property inference, and LLMPhy estimates parameters with a simulator~\citep{tung2023physionpp,cherian2026llmphy}.
These tasks concern conclusions drawn from observations; we examine which observation should come next.

\paragraph{Learning through experiments}
Active perception treats sensing as an action~\citep{bajcsy1988active,aloimonos1988active,bajcsy2018revisiting,bohg2017interactive}.
Learned physical experiments and sequential design seek observations with high information value~\citep{denil2017experiments,foster2021dad}, learned policies can choose which question to ask~\citep{dipta2026decomposerl}, and cost-aware planning scores committed plans against an oracle budget~\citep{nazi2026triage}.
Language models can interleave reasoning with actions or generate executable policies~\citep{yao2023react,liang2023code}, while structured supervision can shape how models learn tool-use patterns and intermediate decisions~\citep{jiang2026scribestructuredmidlevelsupervision,xu2026learningusetoolsjust}.
IVRE, DeepPHY, and HExA study interactive physical reasoning~\citep{xu2023ivre,xu2025deepphy,chandra2026hexa}.
PhysCaP and task sufficient world models further connect exploration to the downstream objective~\citep{lin2026physcap,feng2026tasksufficient}.
We compare pretrained models in a setting where all possible test results and costs are known.

\paragraph{Knowing when to answer}
Selective prediction permits abstention~\citep{geifman2017selective}.
Calibration and uncertainty estimates assess confidence in a response~\citep{guo2017calibration,kadavath2022know,kuhn2023semantic,hossain2026uat}.
Related studies also show that model behavior depends on how reasoning is structured, constrained, distilled, and optimized during post-training~\citep{jiang-etal-2026-drp,jiang-ferraro-2026-beyond,jiang2026bridgingreasoningtrajectoriesonpolicy,yang2026modularizedreinforcementlearningllms}.
TRAPSBench pairs sufficient and insufficient physical evidence to study restraint~\citep{pramono2026trapsbench}, and OMD-Bench corrupts modalities to test calibrated abstention~\citep{nazi2026omni}.
Our task additionally requires a choice among missing measurements.
Complementary images, contrast sets, and behavioral tests motivate comparisons beyond aggregate accuracy~\citep{goyal2017vqa,gardner2020contrast,ribeiro2020checklist,mazumder2026agentcollabbench,sayeedi2026many,lia-roy-dipta-2026-cross}.
Here, the paired construction links each controlled input change to a known change in the correct action.
Robustness is crucial across safety sensitive AI applications, including medical diagnosis and screening~\cite{kamran2019optic,kamran2020comprehensive,saha2022pairwise,ravin2022mitigating,saha2023rfc,sarwar2025fedmentalcare,sarwar2025fedmentor}, accessible navigation and communication~\cite{saha2022mypath,abdullah2025breaking}, traffic sign understanding for autonomous driving~\cite{saha2018total,saha2018efficient}, and reliable visual recognition~\cite{saha2018lightning,saha2025improving}. 
It is equally essential for interpretable vision systems~\cite{saha2023seebel}, multimodal retrieval and reasoning~\cite{saha2026zero,saha2026flipmeasuringhiddenscore,sarwar2025filterrag,roy-dipta-ferraro-2025-q2e,roy-dipta-etal-2026-vc}, generative imaging~\cite{dipta2026oraclezoomonpolicyselfdistillationinspired}, and safe model unlearning and concept erasure~\cite{joshi2024towards,saha2025side,saha2026erase,sarwar2026multimodal}, where brittle predictions or unintended behavior can directly undermine user trust and lead to consequential outcomes. 

%% file: sec/04_method.tex
\section{Method}
\label{sec:method}
\begin{figure}[htbp]
    \centering
    \includegraphics[width=\linewidth]{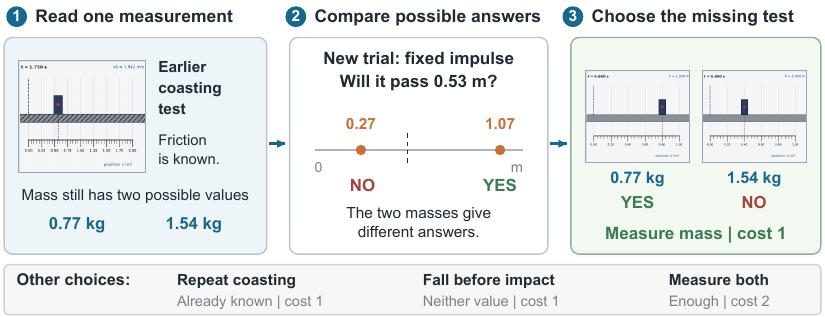}
    \caption{\textbf{One new property can separate the remaining answers.}
    The earlier coasting measurement fixes friction, leaving two masses.
    Their predicted distances lie on opposite sides of the target mark.
    The mass test distinguishes these cases at cost one.
    The bottom row gives the information and prices of the other actions.
    Panel crops come from the same recorded family as Figure~\ref{fig:teaser}.
    }
    \label{fig:method}
\end{figure}

Given an image and a physical question, a model chooses \stopaction or one further experiment.
We first define the candidate worlds, then construct the paired inputs, and finally compute the reference decisions.
Figure~\ref{fig:method} illustrates this process with a recorded sliding problem.

\subsection{Candidate Physical Worlds}
\label{sec:worlds}
A parameter family has four worlds, $\worlds=\{w_{00},w_{01},w_{10},w_{11}\}$.
Each world is one combination of two mass values and two values of another property.
The latter is friction $\mu$, restitution $e$, or spring stiffness $k$.
Restitution describes how strongly an object rebounds.
All four combinations are equally likely before measurement.
We predict sliding distance $d$, energy immediately after a bounce $E$, or oscillation period $T$:
\begin{equation}
    d=\frac{J^2}{2\mu g m^2},\qquad
    E=mge^2h,\qquad
    T=2\pi\sqrt{\frac{m}{k}}.
    \label{eq:physics}
\end{equation}
Here, $J$ is the delivered impulse, $h$ is the drop height, and $g$ is gravity.
The question compares one outcome with a threshold $q$.
Candidate values, calibration constants, and ideal physical assumptions are given in the prompt.
Derivations and sampling ranges appear in Appendix.

\paragraph{Recorded measurements}
Each $672\times672$ image contains four panels with a ruler, marked positions, and times or loads.
Displacement under a known force reveals mass on a frictionless track.
Coasting distance indicates friction, rebound height indicates restitution, and extension under a load indicates stiffness.
The initial image measures one property and leaves two compatible worlds.

\subsection{Paired Inputs}
\label{sec:matched}
We render both possible results of the initial test.
Either image leaves two possible target outcomes.
Write their target outcome ranges as $[a,b]$ and $[c,d]$, ordered by $a<c$ and $b<d$.
Each range contains exactly two discrete possibilities.
We place a threshold halfway inside each interval below:
\begin{equation}
    q_{\mathrm{low}}\in(a,\min(b,c)),\qquad
    q_{\mathrm{high}}\in(\max(b,c),d).
    \label{eq:thresholds}
\end{equation}
The lower threshold leaves the first history unresolved and makes the second imply \yes.
The higher threshold makes the first imply \no and leaves the second unresolved.
This creates two comparisons: exchange the image at a fixed question, or exchange the threshold at a fixed image.
The image comparison preserves all prompt bytes.
The threshold comparison retains the image and action ordering.

\subsection{Reference Actions and Scores}
\label{sec:scoring}
Let $\worlds(H)$ contain worlds compatible with history $H$, and let $y_q(w)$ denote the binary answer in world $w$.
The answer is known when all remaining worlds agree:
\begin{equation}
    R(H,q)=\mathbf{1}\!\left[\left|\{y_q(w):w\in\worlds(H)\}\right|=1\right].
    \label{eq:answerability}
\end{equation}
\Needspace{7\baselineskip}
When $R=1$, stopping costs zero and completes the decision.
Otherwise, action $a$ supplies observation $o_a(w)$ at price $c(a)$.
The reference selects the cheapest action that resolves every remaining case:
\begin{equation}
    a^*\in\argmin_{a\in\actions}c(a)
    \quad\text{subject to}\quad
    R(H\cup\{o_a(w)\},q)=1\quad\forall w\in\worlds(H).
    \label{eq:action}
\end{equation}
A test of either property costs one unit; the combined report costs two.
Free fall before contact costs one and reveals neither unknown.
For an unresolved base question, the missing property is the unique cheapest useful measurement.
Prices are assigned units, not measured robot or sensor costs.

\paragraph{Evaluation measures}
\emph{Minimum cost choice} counts actions matching $a^*$.
\emph{Both correct} requires success on each member of a pair.
\emph{Same choice} compares decoded actions and includes pairs with two invalid replies.
We also score whether the selected evidence settles the question and whether the final answer is right.
\emph{Resolved and correct} requires both conditions, excluding lucky guesses from insufficient observations.
Our objective is to determine whether the available evidence is sufficient to justify an answer, instead of trading off expected answer accuracy against the cost of additional experiments.

%% file: sec/05_experiments.tex
\section{Experiments}
\label{sec:experiments}
We examine changes in action, unnecessary measurements, use of acquired evidence, and the capabilities needed for these decisions.

\paragraph{Data and models}
The study contains $144$ independent families, with $48$ per physical system.
Within each system, half initially measure mass and half measure the other property.
Two histories and two thresholds yield $576$ decisions and $288$ pairs for either comparison.
A separate pilot of $12$ families informed the answer format.
We test Qwen2.5 VL at 3B, 7B, and 32B, SmolVLM2 at 2.2B, Idefics3 at 8B, and Pixtral at 12B~\citep{bai2025qwen25vl,marafioti2025smolvlm,laurencon2024idefics3,agrawal2024pixtral}.
Weights remain unchanged and unquantized.

\paragraph{Generation and coverage}
The direct protocol asks for an option code within $16$ tokens.
The brief protocol appends a request for at most two short sentences and a final \texttt{ANSWER: <option>} line.
Its limit is $256$ tokens, with a recorded maximum of $200$ for Qwen 32B.
Both use greedy decoding and randomized option codes that remain aligned within pairs.
Prompted reasoning and prompt wording can both affect performance~\citep{wei2022cot,kojima2022zeroshot,roy-dipta-ferraro-2025-may,nazi2026dag,islam2026register}; this comparison also changes response length.
An analysis parser decodes the outputs, with invalid replies retained in each denominator.
Some accepted reasoning replies are misread by its fallback rules.

Each direct run contains $9{,}216$ calls, covering selection, answers before and after every test, numerical controls, and perception probes.
Six input variants use a fixed subset of $72$ families.
Qwen 32B has $2{,}304$ brief calls for selection, given parameter questions, ruler reading, and property recovery.
No scores are inferred for its absent conditions.

\paragraph{Study status and uncertainty}
Planned requirements included $75\%$ accuracy for Qwen 7B with all physical properties given.
This and other control requirements failed, but evaluation continued, making the analysis exploratory.
Intervals use $2{,}000$ seeded bootstrap resamples of complete families~\citep{efron1981nonparametric}.
Linked observations remain together, and variant differences use matched subsets.
The $95\%$ intervals are pointwise and do not provide joint coverage of all comparisons.
Appendix records parsing details and deviations from the plan.

%% file: sec/06_results.tex
\section{Results}
\label{sec:results}

\subsection{Different Observations Often Receive the Same Action}
\label{sec:mainresults}
\begin{figure}[htbp]
    \centering
    \includegraphics[width=\linewidth]{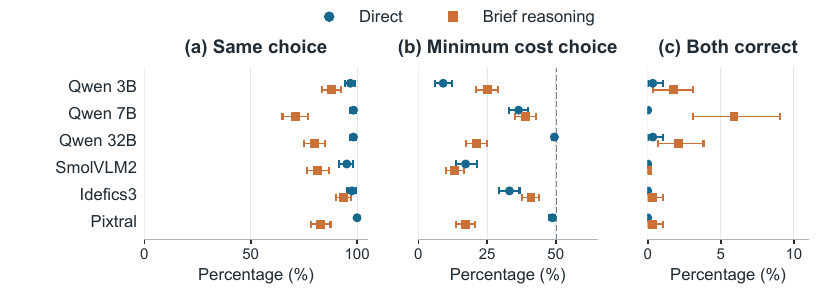}
    \caption{\textbf{Paired scores expose behavior hidden by average accuracy.}
    The panels compare unchanged actions, correct minimum cost decisions, and pairs with both decisions right.
    Each model has $576$ choices and $288$ image pairs per protocol.
    The reference never repeats an action within these pairs; always stopping scores $50\%$ in the middle panel.
    Error bars give $95\%$ family bootstrap intervals.
    }
    \label{fig:decisions}
\end{figure}

Figure~\ref{fig:decisions} shows little adaptation under direct responses.
Across models, at most $0.3\%$ of image pairs have both actions correct.
Changing only the threshold produces repetition rates of $95.5\%$ to $97.9\%$.
Qwen 32B has the highest direct decision score, $49.5\%$, yet almost never solves an entire image pair.
Pixtral stops on $97.6\%$ of decisions and matches no complete image pair.
Appendix separates these measurements by physical system.

\paragraph{Additional working changes the pattern}
Brief reasoning lowers action repetition to between $70.8\%$ and $93.8\%$.
Qwen 7B attains the strongest image pair score, $5.9\%$ [$3.1$, $9.0$].
For Qwen 32B and Pixtral, greater variation accompanies worse decision accuracy.
Some apparent changes involve invalid responses or parser errors, so variation alone cannot establish better evidence use.

\begin{table}[t]
    \centering
    \caption{\textbf{Extra reasoning helps one pair and leaves another error unchanged.}
    Qwen 7B responses are decoded from explicit final letters.
    The required row gives the cheapest correct action; plus and cross mark correctness.
    These examples were selected after evaluation.
    Crops show the last panel, although the model received each full sheet.
    }
    \label{tab:qualitative}
    \includegraphics[width=\linewidth]{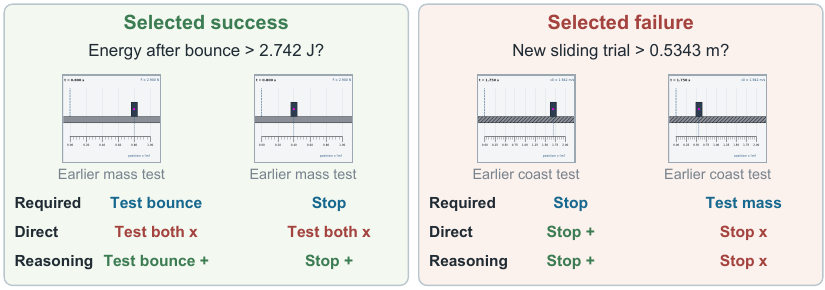}
\end{table}

Table~\ref{tab:qualitative} provides concrete cases behind the protocol comparison.
The bounce example contains a correct switch after brief working.
Table~\ref{tab:explanations} contrasts the accompanying text with the physical setup.
A correct action on the first sliding history does not validate its explanation.

\begin{table}[htbp]
    \centering
    \caption{\textbf{Written explanations can name the missing property or confuse the trial.}
    Exact excerpts from Qwen 7B's brief responses in Table~\ref{tab:qualitative} are shown beside our reading of them.
    The excerpts describe generated text, not access to internal reasoning.
    }
    \label{tab:explanations}
    \small
    \setlength{\tabcolsep}{4pt}
    \begin{tabular}{@{}p{0.11\linewidth}p{0.45\linewidth}p{0.38\linewidth}@{}}
    \toprule
    Case & Recorded excerpt & Interpretation \\
    \midrule
    Bounce & ``provides information about the object's mass but not the coefficient of restitution.'' & The response identifies which of the two properties is absent. \\
    \addlinespace
    Sliding & ``provides direct evidence of the block's stopping distance.'' & The image starts at a known speed. The target starts after an impulse. \\
    \bottomrule
    \end{tabular}
\end{table}

\subsection{Obtaining Evidence and Using It Can Fail Separately}
\label{sec:repeated}
\label{sec:answers}
\begin{table}[t]
    \centering
    \caption{\textbf{Direct results separate evidence, answers, and repeated tests.}
    Direct scores are percentages with $95\%$ intervals.
    Evidence and final success use $576$ decisions; supplied test accuracy uses $1{,}152$ answers after informative measurements on unresolved cases.
    The repeat rate includes only unresolved cases where the model buys a property test; $n$ gives that count.
    }
    \label{tab:main-evidence}
    \input{tab/main_evidence}
\end{table}

Table~\ref{tab:main-evidence} distinguishes available information from the answer it produces.
The selected evidence is sufficient in at least half the decisions, partly because those questions require no new observation.
With useful tests supplied on unresolved cases, answer accuracy reaches only $34.6\%$ at best.
Qwen 7B, Qwen 32B, Idefics3, and Pixtral instead give \unknown on almost every direct request after a test.

\paragraph{Measuring the known property}
Among the $278$ single property purchases made by Qwen 3B on unresolved cases, $83.5\%$ repeat the earlier test.
SmolVLM2 and Idefics3 show a similar preference.
Qwen 32B rarely makes this particular error, but stops prematurely on $75.7\%$ of unresolved questions.
These outcomes separate choosing the right measurement from deciding whether to measure.
The three Qwen sizes do not establish a scaling law.

\begin{table}[t]
    \centering
    \caption{\textbf{One model repeats a test; another cannot use the useful result.}
    The two direct response cases show the earlier image, selected action, and answers obtained with the needed test supplied.
    Blue gives each possible outcome and its correct answer; the line below shows the model response.
    Labels translate recorded option letters, not quoted explanations.
    The rows illustrate contrasting failures found in the completed records, without estimating their frequency.
    Each image is one panel from a full input sheet.
    }
    \label{tab:qualitative-extra}
    \includegraphics[width=\linewidth]{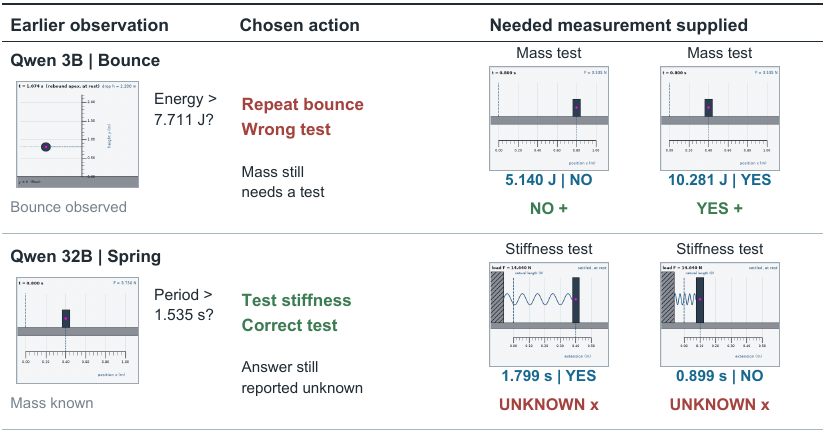}
\end{table}

The additional examples in Table~\ref{tab:qualitative-extra} make this separation explicit.
Qwen 3B requests another bounce test, yet succeeds for both masses when their measurements are supplied.
Qwen 32B selects stiffness correctly in the spring problem, then withholds a definite answer despite sufficient information.
The displayed answers come from the saved requests for those two useful measurements.

\paragraph{Changing the visible measurement}
On the fixed subset, Qwen 3B repeats the base test on $79.9\%$ of property choices with the original image, $83.5\%$ with a new rendering, and $51.1\%$ without an image.
Comparable shifts occur for SmolVLM2 and Idefics3.
Deleting the image changes both available evidence and the conditional choice population.
The correct action is therefore recomputed; this comparison does not isolate labels from geometry.

\subsection{The Capability Controls Remain Weak}
\label{sec:controls}
\begin{table}[t]
    \centering
    \caption{\textbf{Reading and calculation still matter when selection is removed.}
    Values are percentages with pointwise $95\%$ intervals.
    Ruler reading allows $5\%$ relative error; property recovery has two options.
    These probes each use $288$ items.
    Both given means that all property values are supplied as numbers on $1{,}152$ questions.
    Accepted gives the percentage parsed into an answer in that direct condition.
    }
    \label{tab:main-controls}
    \input{tab/main_controls}
\end{table}

Visual judgments can fail before a model chooses an experiment~\citep{rahmanzadehgervi2024blind,tong2024eyes,li2023pope}.
Table~\ref{tab:main-controls} shows that Qwen 7B often locates the ruler mark but struggles to recover the measured property.
Supplying both properties removes image reading from the task and tests numerical mechanics~\citep{ding2023physqa}.
No model reaches the planned $75\%$ accuracy requirement in this condition.
Thus, the main findings cannot be isolated from broader physical reasoning failures.

\paragraph{A low score can reflect an unfinished response}
Pixtral's direct given parameter accuracy is $0.1\%$, with only one accepted answer among $1{,}152$ requests.
Many generations start a derivation and hit the token cap.
Brief responses reach $45.5\%$, showing why the original score should not be read as near total lack of physics knowledge; recoverable knowledge can be missing from a default response~\citep{hossain2026rightknowledgewronganswer}.
A separate early stopping rule can also truncate numerical ruler responses before their final value.
Appendix provides validity across all request types.

\paragraph{Prompt changes offer no common remedy}
Text readings improve direct choice accuracy by $9.0$ to $14.2$ percentage points for four models.
They leave Qwen 32B unchanged and lower Pixtral by $35.8$ points, with $45.5\%$ valid replies in that condition.
Asking models to check the remaining possibilities improves SmolVLM2 by $10.1$ points [$5.6$, $14.9$].
Giving equations adds $5.2$ points [$1.7$, $9.0$] for the same model, while all five other equation intervals include zero.
Table~\ref{tab:main-variants} compares these changes across all six models; Appendix adds the remaining variants.

\begin{table}[htbp]
    \centering
    \caption{\textbf{The same prompt change can help one model and harm another.}
    Entries give changes in direct minimum cost accuracy, in percentage points, relative to the matching $72$ family base subset.
    Brackets show paired $95\%$ intervals.
    Colors and style change together in the new rendering; the remaining possibilities instruction adds no physical evidence.
    }
    \label{tab:main-variants}
    \input{tab/main_variants}
\end{table}

\subsection{The Selection Rule Also Changes Cost}
\begin{table}[htbp]
    \centering
    \caption{\textbf{A cheaper sufficient measurement can improve final success.}
    The rules recombine direct answer records over $576$ decisions, without further model calls.
    Each model reads the evidence under all three rules.
    Success means resolved and correct; cost is in assigned units.
    Brackets are $95\%$ intervals.
    The minimum cost reference optimizes sufficiency, not the accuracy of a fallible reader.
    }
    \label{tab:main-cost}
    \input{tab/main_cost}
\end{table}

Table~\ref{tab:main-cost} compares model choices with two reference rules.
For Qwen 3B, selecting the cheapest sufficient evidence gives $31.7\%$ final success at cost $0.50$, compared with $18.8\%$ at $0.99$ for its own choices.
SmolVLM2 also has a higher point estimate at lower cost under this rule.
The combined report costs four times as much as the minimal rule, with no gain shared by both models.
Appendix includes all six models and alternative report prices.
Those price changes rescore stored actions; they do not test a model's response to a newly advertised price.

%% file: sec/07_conclusion.tex
\section{Conclusion}
We introduced a paired evaluation for testing whether vision language models know when they have enough evidence to answer a physical question and, when they do not, which additional experiment they should perform.
Because each pair is designed so that a controlled change in the observation or question changes the correct action, every model choice can be checked against an exact reference.
Across six open models, aggregate scores hide a central weakness: models often repeat the same action even when new evidence should change their decision.
Our analysis also shows that selecting a useful experiment and correctly interpreting its result are separate challenges.
Future progress requires models that adapt their choices to the available evidence while also reading measurements, performing calculations, and formatting answers reliably.

%% file: sec/08_limitations.tex
\section{Limitations}
\label{sec:limitations}
Our findings cover specific open model checkpoints, three idealized systems, discrete disclosed properties, and at most one additional experiment.
We do not test natural video, sensor noise, physical robots, or open ended experiment design.
Selection and answering use separate calls, so the answering model never sees the selection rationale.
Some analyses followed output inspection, and several Qwen 32B brief reasoning conditions are missing.
Image ablation and extended generation alter multiple factors, limiting causal attribution.
The deterministic reference receives additional metadata and images; its perfect score validates the procedure but is not directly comparable.
Thus, our results establish systematic behavioral failures without identifying a unique internal cause.

%% file: sec/09_appendix.tex
\section{Physical Systems and Matched Construction}
\label{sec:construction}
This appendix gives the construction details and complete results.
Qwen 3B, Qwen 7B, and Qwen 32B refer to the three Qwen checkpoints in Appendix.
SmolVLM2, Idefics3, and Pixtral denote the checkpoints named 2.2B, 8B, and 12B, respectively.
\emph{Direct} requests an immediate option code.
\emph{Reason} denotes the brief reasoning protocol used in the main tables.
Unless stated otherwise, values are percentages with $95\%$ bootstrap intervals in brackets.
Each resample contains complete parameter families.
NA marks measurements that are not available.

\subsection{Physical Assumptions and Derivations}
\paragraph{Sliding}
An impulse $J$ changes a block's velocity from zero to $v_0=J/m$.
Coulomb friction produces deceleration $\mu g$ until the block stops.
Using $0=v_0^2-2\mu g d$ gives the first expression in Equation~\ref{eq:physics}.
The block slides without rolling on a level surface.
The coefficient is constant, no drag acts, and no obstacle interrupts the motion.
The coasting measurement uses the same surface and starts at a known speed.
The target trial starts with a known impulse, so its initial speed depends on mass.

\paragraph{Bouncing}
An object dropped from height $h$ has impact speed $\sqrt{2gh}$.
Its rebound speed is $e\sqrt{2gh}$.
The kinetic energy immediately after impact is therefore $\frac12m(e\sqrt{2gh})^2=mge^2h$.
Motion is vertical, with no rotation or drag.
The floor is fixed and the restitution coefficient is constant over the tested impact speeds.
The measurement and target trial use the same object and floor.

\paragraph{Spring motion}
The displacement $x(t)$ of a horizontal mass on a linear spring obeys $m\ddot{x}=-kx$.
The angular frequency is $\sqrt{k/m}$, giving $T=2\pi\sqrt{m/k}$.
The spring is massless, friction and damping are negligible, and motion stays in the linear regime.
The stiffness measurement uses the same spring under a known static load.

\subsection{Parameter Sampling and Calibration}
\begin{table}[htbp]
    \centering
    \caption{\textbf{Each family varies two disclosed physical properties.}
    Lower values are sampled independently within the shown ranges.
    The upper value is a fixed multiple of the lower value.
    The generator rounds stored physical quantities to six significant digits.
    }
    \label{tab:ranges}
    \small
    \begin{tabular}{llcl}
    \toprule
    Quantity & Distribution of lower value & Upper multiplier & Units \\
    \midrule
    Mass $m$ & Uniform $[0.75,1.50]$ & $2$ & kg \\
    Friction $\mu$ & Uniform $[0.08,0.12]$ & $3$ & dimensionless \\
    Restitution $e$ & Uniform $[0.30,0.45]$ & $2$ & dimensionless \\
    Stiffness $k$ & Uniform $[20,40]$ & $4$ & N/m \\
    Sliding impulse $J$ & Uniform $[1.5,2.5]$ & NA & N\,s \\
    Target drop height $h$ & Uniform $[0.6,1.4]$ & NA & m \\
    Gravity $g$ & $9.81$ & NA & m/s$^2$ \\
    \bottomrule
    \end{tabular}
\end{table}
The main generator uses seed $20260905$.
It creates $48$ families per system.
Within each system, it randomly orders $24$ initial mass tests and $24$ initial tests of the other property.
The diagnostic subset uses seed $20260906$.
It selects $12$ families for each combination of system and initial test, giving $72$ families.
The independent pilot contains $12$ further families.
It is excluded from the reported main results.

Mass calibration uses a frictionless $1$\,m track and timestamps $0.2,0.4,0.6,0.8$\,s.
The disclosed force is chosen so the lower mass reaches $0.8$\,m in the final panel.
The resulting displacement is $x=Ft^2/(2m)$.
The coasting test uses a $2$\,m track.
Its known launch speed makes the case with lower friction stop at $1.7$\,m.
The four timestamps divide that stopping time into quarters.
The rebound test chooses its drop height so the case with higher restitution reaches $0.8$\,m on a $1$\,m ruler.
The spring test applies four loads, with the maximum extending the softer spring by $0.4$\,m on a $0.5$\,m ruler.
Free fall begins at $1.2$\,m and ends at $0.2$\,m above the floor.
Calibration constants depend only on the disclosed candidate values.
They do not depend on which world is hidden.

\subsection{Why the Required Action Changes}
Consider the ordered outcome ranges used in Equation~\ref{eq:thresholds}.
Since $a<q_{\mathrm{low}}<\min(b,c)$, the first history admits outcomes on both sides of the lower threshold.
Both outcomes of the second history exceed this threshold.
For the higher threshold, $\max(b,c)<q_{\mathrm{high}}<d$ puts both outcomes of the first history below the threshold and splits the second history.
Each history therefore switches between a known answer and an unknown answer across the two questions.
Each question also switches between those states across the two histories.
Applying the action rule in Section~\ref{sec:scoring} then gives the stated switch at each threshold.

\paragraph{Indistinguishable hidden alternatives}
Suppose two equally likely remaining worlds produce the same observation and opposite binary answers.
An answer rule restricted to \yes and \no gives the same output distribution in both worlds when it uses only this observation.
If it answers \yes with probability $p$, its mean binary correctness across the two worlds is $(p+(1-p))/2=1/2$.
An additional measurement with the same outcome in both worlds leaves this bound unchanged.
This bound explains why a correct guess does not count as evidence that settles the answer.
It follows from the construction and is not a new impossibility result.

\subsection{Checks on the Rendered Evidence}
\begin{table}[htbp]
    \centering
    \caption{\textbf{The stored construction passes its implemented checks.}
    Pixel checks use the original rendered sheets.
    They do not test every model's resized or tokenized visual input.
    }
    \label{tab:construction}
    \small
    \begin{tabular}{lr}
    \toprule
    Check & Recorded result \\
    \midrule
    Independent parameter families & $144$ \\
    Candidate worlds & $576$ \\
    Initial histories / decisions & $288$ / $576$ \\
    Required switches after image / threshold change & $288$ / $288$ \\
    Checks for identical text passed & $288/288$ \\
    Checks for identical images across hidden worlds passed & $1{,}152/1{,}152$ \\
    Largest error in extracting the marked position & $0.456\%$ of panel span \\
    Smallest informative feature separation & $54$ pixels \\
    Required separation in implementation & $8$ pixels \\
    Smallest relative threshold margin & $17.2\%$ \\
    Reported check failures & $0$ \\
    \bottomrule
    \end{tabular}
\end{table}
The measurement renderer marks the relevant position with a magenta dot and a line to the ruler.
A deterministic detector compares this position with the generating value.
Image hashes verify that measurements which cannot distinguish hidden alternatives reuse identical bytes.
The text checks cover each history pair with a fixed question.
The rendering checks support the intended geometric encoding.
They do not establish that each learned model reads it correctly.

\paragraph{Scope of the deterministic reference}
The stored deterministic program obtains a correct result on all $9{,}216$ of its records.
It uses known calibration values, panel bounds, candidate metadata, and item identifiers.
For combined reports, it reads the two separate measurement sheets.
For the text reading variant, it reads the original image.
For questions with both parameter values given, it obtains the world from the item identifier.
Its score checks the implementation.
The program receives information unavailable to the models, so its score cannot rule out errors in how they read the images.

\Needspace{13\baselineskip}
\section{Recorded Protocol and Reproducibility Details}
\label{sec:protocol}
\subsection{Inference Accounting}
\begin{table}[H]
    \centering
    \caption{\textbf{All direct runs cover the full item set.}
    Five reasoning runs have the same coverage.
    Qwen 32B reasoning covers selection and the three control groups.
    }
    \label{tab:coverage}
    \small
    \begin{tabular}{lrr}
    \toprule
    Request type & Full run & Qwen 32B reasoning \\
    \midrule
    Initial action selection & $576$ & $576$ \\
    Answer from initial evidence & $576$ & NA \\
    Answers after four tests in both hidden worlds & $4{,}608$ & NA \\
    Full numerical state & $1{,}152$ & $1{,}152$ \\
    Read the marked position & $288$ & $288$ \\
    Recover the measured property & $288$ & $288$ \\
    Six input variants & $1{,}728$ & NA \\
    \midrule
    Total & $9{,}216$ & $2{,}304$ \\
    \bottomrule
    \end{tabular}
\end{table}
The direct and reasoning runs contain $55{,}296$ and $48{,}384$ model calls, respectively.
The main study therefore contains $103{,}680$ calls.
The saved completion checks find no duplicate item identifiers or malformed record lines.
All completed VLM records have successful infrastructure status.
That status is separate from a valid or correct generated answer.
The deterministic program's $9{,}216$ records are additional reference computations, not VLM calls.

\subsection{Output Generation and Parsing}
The option menu maps five letters to \stopaction and the four experiments.
The action ordering is sampled per family and reused across both histories and questions.
Answer requests use a randomized menu with three options: \yes, \no, and \unknown.
Its ordering is preserved across the matched histories.
The pilot used bare answer words.
Its frequent constant answers led to the lettered format.

The reasoning request is appended to the original prompt, which still includes its request for only an option letter.
The exact appended text for an option response is:
\begin{quote}\small
Give at most two short sentences of working. You must then end your reply with this exact final line and nothing after it:\\
\texttt{ANSWER: <option letter>}
\end{quote}
This suffix defines the protocol contrast in Section~\ref{sec:experiments}.
The implementation defaults to a maximum of $256$ tokens.
Five runs reach that limit; Qwen 32B reaches $200$ tokens.
The direct cap is $16$ tokens.

All reported metrics use the saved analysis parser.
For reasoning, it first seeks a final option letter, then tries standalone option letters in the response.
Answer words and explicit action names provide fallbacks.
This rule can read a letter in the explanation as the chosen option when the final answer is malformed.
For example, a Qwen 7B sliding reply ends in \texttt{ANSWER: STOP}.
The parser scores it as a mass measurement because it matches an earlier standalone \texttt{D}.
Another truncated bounce reply names the test of the second property but is scored as mass.
We retain the supplied metrics.
A reasoning answer counted as correct by this parser may not express the correct action.
The qualitative records are identified separately in Appendix.

For numerical ruler responses, the reasoning stopping rule includes the prefix \texttt{ANSWER:} itself.
This can end generation before the final number.
The recorded numeric parser can then obtain a number from the preceding explanation.
Consequently, reasoning ruler scores combine measurement reading, output formatting, and extraction behavior.
Direct ruler error is the absolute reading error divided by the true reading, not the full ruler span.
This differs from the renderer's geometric check in Table~\ref{tab:construction}.

\paragraph{Interpreting unchanged choices}
The unchanged choice rate compares decoded actions.
Two invalid actions are also equal in the implementation.
A pair can stop counting as unchanged when only one member becomes invalid.
SmolVLM2's reasoning selection validity is $78.0\%$; the other models range from $99.8\%$ to $100\%$.
The direct selection validity is $100\%$ for all six models.
This issue affects the reasoning comparison; direct selection has no invalid actions.

\subsection{Controls and Changes from the Study Plan}
The plan required Qwen 7B to reach $75\%$ answer accuracy when both parameters were given.
It also required Qwen 7B and SmolVLM2 to reach $80\%$ measurement reading accuracy.
The output validity requirement was $95\%$ on applicable requests.
The executed direct study did not satisfy all of these requirements.
Qwen 7B scores $49.6\%$ with both parameters given, and SmolVLM2 scores $42.0\%$ on ruler reading.
Some output groups also fall below the validity requirement.
The study nevertheless continued.
These requirements appear in the research plan.
The supplied record does not document an independent public registration.

The final study also adds three model checkpoints, the two full perception probes, the deterministic reference program, the reasoning protocol, and additional behavioral summaries.
Parsing by action name was added after inspecting responses.
The 32B reasoning run was shortened because its full projected runtime exceeded the available window.
The 7B reasoning run was resumed to complete its remaining requests.
These changes support the exploratory scope stated in Section~\ref{sec:experiments}.
The result tables retain every model and physical system.

\subsection{Uncertainty and Reporting}
The recorded bootstrap uses $2{,}000$ resamples with seed $20260905$.
It resamples the $144$ families, preserving all related measurements and questions, or the $72$ families for a variant comparison.
It does not treat repeated deterministic responses as independent trials.
The reported intervals describe variation across these sampled parameter families under one recorded run configuration.
They do not measure variation across checkpoints, decoding seeds, or data sources.
An interval such as $0[0,0]$ is a degenerate empirical bootstrap result, not certainty that the population rate is zero.

Some stored significance values can exceed one when every difference is zero.
The implementation also omits the monotonic adjustment required by the standard Holm procedure.
We exclude these values and make no adjusted significance claims.
We report point estimates and pointwise percentile intervals, with no simultaneous coverage claim.
Protocol contrasts are descriptive unless a paired interval is explicitly provided.
Variant tables include the stored paired difference intervals.
These intervals do not account for parser errors.

\section{Complete Decision Results}
\label{sec:fullresults}
The next tables give all choice scores, pair scores, and unchanged choice rates.
Both members of a pair must match the reference action to count as correct.
The individual score uses $576$ decisions; each pair score uses $288$ pairs.
Each system table uses $48$ families and $192$ decisions.
They do not show a consistent ranking of physical systems across all models.

\newcommand{\resulttable}[3]{%
\begin{table}[htbp]
    \centering
    \caption{#2}
    \label{#3}
    \input{tab/#1}
\end{table}}

\begin{table}[htbp]
    \centering
    \caption{\textbf{Choice accuracy separates stopping from measuring.} ``Stop correctly'' uses the $288$ questions with a known answer. ``Measure correctly'' uses the $288$ unresolved questions.}
    \label{tab:all-decisions}
    \input{tab/r1_decisions}
\end{table}
\begin{table}[htbp]
    \centering
    \caption{\textbf{Both choices in a pair are seldom correct.} Changing the image preserves all prompt text. Changing the threshold preserves the image and action menu.}
    \label{tab:pairs}
    \input{tab/r1_pair_correctness}
\end{table}
\begin{table}[htbp]
    \centering
    \caption{\textbf{Most pairs retain the same decoded action.} The reference rate is zero. A change alone does not imply correctness. Appendix explains the effect of invalid outputs.}
    \label{tab:same-actions}
    \input{tab/r1_repeated_actions}
\end{table}
\FloatBarrier

\begin{table}[htbp]
    \centering
    \caption{\textbf{Sliding results.} Minimum cost choice accuracy is reported under each response protocol.}
    \label{tab:system-slide}
    \input{tab/r1_system_slide}
\end{table}
\begin{table}[htbp]
    \centering
    \caption{\textbf{Bouncing results.} The target asks whether rebound energy exceeds the stated threshold.}
    \label{tab:system-bounce}
    \input{tab/r1_system_bounce}
\end{table}
\begin{table}[htbp]
    \centering
    \caption{\textbf{Spring results.} The target asks whether the period exceeds the stated threshold.}
    \label{tab:system-spring}
    \input{tab/r1_system_spring}
\end{table}
\FloatBarrier

\section{Evidence Selection and Answering}
\label{sec:fullanswers}
Let $S$ mean that the selected evidence settles the answer, and let $B$ mean that the final answer is correct.
Then $P(S\cap B)=P(S)P(B\mid S)$ for the same set of policy decisions.
The supplied answer accuracy uses a different set.
It covers the missing property test and combined report on initially unresolved questions, in both hidden worlds.
Each complete run has $1{,}152$ such answers.
This score is not $P(B\mid S)$ for the model's selected actions.
Table~\ref{tab:evidence-answer} keeps the three measurements separate.

\begin{table}[htbp]
    \centering
    \caption{\textbf{Complete evidence and answer scores for both protocols.} Selection sufficiency uses $576$ decisions. Answer accuracy uses the $1{,}152$ informative measurement answers defined above. Final success averages both hidden worlds for each decision. Qwen 32B reasoning answers are unavailable.}
    \label{tab:evidence-answer}
    \input{tab/r2_evidence_and_answers}
\end{table}
\begin{table}[htbp]
    \centering
    \caption{\textbf{An unknown answer does not always lead to a new measurement.} ``Known answer'' scores the $288$ initially settled questions. ``Unknown answer'' scores the $288$ unresolved questions. ``Action agreement'' compares saying undetermined with buying a test across all $576$ decisions. Agreement alone is not accuracy.}
    \label{tab:initial-answers}
    \input{tab/r2_initial_answers}
\end{table}
\paragraph{Repeated unknown answers}
Several models answer \unknown on almost every direct request after a measurement.
Out of $4{,}608$ answers per model, the counts are $4{,}537$ for Qwen 7B, $4{,}582$ for Qwen 32B, $4{,}595$ for Idefics3, and all $4{,}608$ for Pixtral.
These totals include measurements that do and do not settle the answer.
This repeated answer helps explain their accuracy near zero when the evidence is sufficient.
It is a property of the recorded answers, not proof that every numerical calculation failed.

\paragraph{Hidden alternatives}
The direct run contains $576$ pairs where an extra measurement cannot distinguish the two hidden worlds.
All six models give identical decoded responses to the identical inputs in these pairs.
When they produce a definite binary answer, the average correctness over the balanced hidden alternatives is exactly $50\%$.
Pixtral never produces a definite answer in this group, so its conditional binary rate is undefined.
This matches the construction's expected behavior.
\FloatBarrier

\section{Which Experiments Are Chosen?}
\label{sec:fullchoices}
The next tables separate questions with known and unknown answers.
Each row contains $288$ choices.
``Repeat'' measures the already observed property.
``Missing'' measures the other property.
``Both'' buys the combined report.
Free fall ends before floor contact and reveals neither property.

\begin{table}[htbp]
    \centering
    \caption{\textbf{Choices when the answer is already known.} Only stopping has minimum cost in this group.}
    \label{tab:counts-known}
    \input{tab/r3_action_counts_determined}
\end{table}
\begin{table}[htbp]
    \centering
    \caption{\textbf{Choices when the answer remains unknown.} Measuring the missing property costs less than buying both measurements.}
    \label{tab:counts-unknown}
    \input{tab/r3_action_counts_unresolved}
\end{table}
\begin{table}[htbp]
    \centering
    \caption{\textbf{Unnecessary measurement and early stopping are different errors.} ``Unneeded test'' uses initially determined cases. ``Useless test'' means repeating the known property or selecting free fall on unresolved cases. ``Early stop'' also uses unresolved cases. Each condition has $288$ decisions.}
    \label{tab:wasted}
    \input{tab/r3_wasted_choices}
\end{table}
\FloatBarrier

\begin{table}[htbp]
    \centering
    \caption{\textbf{Several models favor measuring the known property again.} Rates include only choices of mass or the other property test. Counts give the denominators. Uniform selection between these two tests gives $50\%$.}
    \label{tab:repeat-all}
    \input{tab/r3_repeated_test_all}
\end{table}
\begin{table}[htbp]
    \centering
    \caption{\textbf{Repeating the known property leaves unresolved questions unanswered.} The same conditional rate is restricted to initially unresolved questions. Denominators differ from Table~\ref{tab:repeat-all}; Pixtral's direct rate is based on only seven choices.}
    \label{tab:repeat-unknown}
    \input{tab/r3_repeated_test_unresolved}
\end{table}
\FloatBarrier

\section{Measurement Cost and Final Answers}
\label{sec:policies}
Policy comparisons reuse the saved answers after every measurement.
For each initial history and question, scores average the two hidden alternatives.
We look up each action's answer from an independently recorded response.
``Missing property'' always measures the property absent from the initial history, including cases where that measurement is unnecessary.
``Uniform test'' averages exactly over all four measurements, excluding stop.
Its assigned mean cost is $1.25$ and the recorded analysis supplies no interval.
The exact selector stops when possible and otherwise obtains the cheapest sufficient evidence.
Its mean cost is $0.5$ in the balanced base task.
An exact reader with the model selector is a diagnostic bound on evidence sufficiency, not a learned system.

Table~\ref{tab:main-cost} presents the main cost comparison.
\begin{table}[htbp]
    \centering
    \caption{\textbf{Qwen 3B policy comparison.} Final success requires evidence that settles the question and a correct answer. Mean cost includes failed decisions.}
    \label{tab:policy-qwen3}
    \input{tab/r4_policies_qwen3}
\end{table}
\begin{table}[htbp]
    \centering
    \caption{\textbf{Qwen 7B policy comparison.} All policies share the same answer records within each response protocol.}
    \label{tab:policy-qwen7}
    \input{tab/r4_policies_qwen7}
\end{table}
\FloatBarrier
\begin{table}[htbp]
    \centering
    \caption{\textbf{Qwen 32B policy comparison.} Reasoning answers are absent. The $576$ recorded choices still give costs and the evidence score with an exact reader.}
    \label{tab:policy-qwen32}
    \input{tab/r4_policies_qwen32}
\end{table}
\begin{table}[htbp]
    \centering
    \caption{\textbf{SmolVLM2 policy comparison.} Mean cost and final success describe different requirements.}
    \label{tab:policy-smol}
    \input{tab/r4_policies_smol}
\end{table}
\begin{table}[htbp]
    \centering
    \caption{\textbf{Idefics3 policy comparison.} Answer scores near zero can occur even when the selected measurement is informative.}
    \label{tab:policy-idefics}
    \input{tab/r4_policies_idefics}
\end{table}
\begin{table}[htbp]
    \centering
    \caption{\textbf{Pixtral policy comparison.} The model selector is compared with fixed tests and an exact selector.}
    \label{tab:policy-pixtral}
    \input{tab/r4_policies_pixtral}
\end{table}
\FloatBarrier

\subsection{Repricing the Combined Report}
We hold the saved actions fixed and change the combined report price to $1$, $2$, $3$, or $5$.
At price one, the missing property test and combined report are equally cheap ways to settle unresolved questions.
The scorer gives credit for this tie.
Every recorded choice comes from a prompt with the original price of two.
The procedure changes only the scorer's cost lookup.
For example, Qwen 7B's direct choice score rises from $36.5\%$ to $46.5\%$ at price one.
Its combined report choices now receive credit when the answer is unresolved.

\begin{table}[htbp]
    \centering
    \caption{\textbf{Direct choice accuracy after repricing.} Each column scores the same actions at a different combined report price.}
    \label{tab:reprice-direct}
    \input{tab/r4_reprice_accuracy_direct}
\end{table}
\begin{table}[htbp]
    \centering
    \caption{\textbf{Direct mean costs after repricing.} Failed decisions remain in every denominator. Other action prices stay fixed.}
    \label{tab:reprice-cost-direct}
    \input{tab/r4_reprice_cost_direct}
\end{table}
\begin{table}[htbp]
    \centering
    \caption{\textbf{Reasoning choice accuracy after repricing.} All six models have complete selection records, including Qwen 32B.}
    \label{tab:reprice-reason}
    \input{tab/r4_reprice_accuracy_reason}
\end{table}
\begin{table}[htbp]
    \centering
    \caption{\textbf{Reasoning mean costs after repricing.} Costs use the saved decoded actions.}
    \label{tab:reprice-cost-reason}
    \input{tab/r4_reprice_cost_reason}
\end{table}
\FloatBarrier

\Needspace{8\baselineskip}
\section{Perception, Calculation, and Output Validity}
\label{sec:fullcontrols}
The ruler task asks for the last panel's marked reading.
It passes when the absolute error is at most $5\%$ of the true reading.
The second task asks which of two candidate physical values explains that measurement.
Uniform choice between these two values gives $50\%$.
The calculation control gives both physical properties and asks the target question.
Its true binary answers are balanced, so always answering one binary class obtains $50\%$.
The menu also permits undetermined, but no question in this control requires it.

\begin{table}[htbp]
    \centering
    \caption{\textbf{Neither protocol reaches the planned calculation accuracy.} Ruler reading and property recovery each use $288$ probes per model. The given parameters control uses $1{,}152$ questions. Generation and parsing affect all scores.}
    \label{tab:full-controls}
    \input{tab/r5_controls}
\end{table}
\begin{table}[htbp]
    \centering
    \caption{\textbf{Direct selections are valid, but some control outputs are heavily truncated.} Values give the percentage accepted by the parser. A valid answer can still be wrong.}
    \label{tab:valid-direct}
    \input{tab/r5_validity_direct}
\end{table}
\begin{table}[htbp]
    \centering
    \caption{\textbf{Longer responses still have format errors.} Missing Qwen 32B groups are marked NA. An accepted response may still be interpreted incorrectly by the parser.}
    \label{tab:valid-reason}
    \input{tab/r5_validity_reason}
\end{table}
\FloatBarrier

\section{All Input and Prompt Variants}
\label{sec:variants}
Each variant has $288$ decisions from the same fixed $72$ families.
Its difference is computed against the matched base subset within the same response protocol.
The main paper reports changes for direct responses.
Here we also include all five complete reasoning runs.
Qwen 32B reasoning variants were not run.
Changes are in percentage points, with paired intervals from resampling families.

\begin{table}[htbp]
    \centering
    \caption{\textbf{Base scores for the fixed variant subset.} Every variant is compared with this subset, not the full $144$ families.}
    \label{tab:variant-base}
    \input{tab/r6_base}
\end{table}
\begin{table}[htbp]
    \centering
    \caption{\textbf{Text readings have mixed effects.} The text gives measured coordinates and times or loads. It does not reveal the hidden property. Pixtral's direct output validity is $45.5\%$ here.}
    \label{tab:variant-readings}
    \input{tab/r6_variant_readings}
\end{table}
\begin{table}[htbp]
    \centering
    \caption{\textbf{Changing the rendering preserves the measurement values.} Colors and drawing style change together. Small intervals containing zero do not establish invariance to every possible rendering.}
    \label{tab:variant-rendering}
    \input{tab/r6_variant_altrender}
\end{table}
\begin{table}[htbp]
    \centering
    \caption{\textbf{Wording and option order change together.} Their separate effects cannot be inferred from this variant.}
    \label{tab:variant-wording}
    \input{tab/r6_variant_paraphrase}
\end{table}
\begin{table}[htbp]
    \centering
    \caption{\textbf{Supplying the mechanics equations is not a common fix across models.} The added text includes both measurement and target equations.}
    \label{tab:variant-equations}
    \input{tab/r6_variant_equations}
\end{table}
\begin{table}[htbp]
    \centering
    \caption{\textbf{Asking models to check remaining possibilities helps some settings and harms others.} The instruction adds no new physical evidence or equations.}
    \label{tab:variant-remaining}
    \input{tab/r6_variant_remaining}
\end{table}
\begin{table}[htbp]
    \centering
    \caption{\textbf{Removing the measurement changes what must be learned.} All four worlds are now possible, and every question needs the combined report. Labels are recomputed for this condition. The difference is not an input deletion with unchanged correct actions.}
    \label{tab:variant-noimage}
    \input{tab/r6_variant_noimage}
\end{table}
\FloatBarrier

\subsection{Repeated Measurements Under Each Variant}
These rates include only choices to measure one property.
The selected population and its size change across variants.
Without the image, ``repeat'' means choosing the property measured in the matched base case.
The model does not see that measurement in this variant.
Every rate must therefore be read with its denominator.
Section~\ref{sec:repeated} compares the base, changed rendering, and removed image conditions.

\begin{table}[htbp]
    \centering
    \caption{\textbf{Repeated measurements with text readings.} Percentages and counts are restricted to choices of one property test.}
    \label{tab:repeat-readings}
    \input{tab/r6_repeat_readings}
\end{table}
\begin{table}[htbp]
    \centering
    \caption{\textbf{Repeated measurements with changed rendering.} The physical values are unchanged.}
    \label{tab:repeat-rendering}
    \input{tab/r6_repeat_altrender}
\end{table}
\begin{table}[htbp]
    \centering
    \caption{\textbf{Repeated measurements with changed wording and option order.} The selected group can differ substantially from the base condition.}
    \label{tab:repeat-wording}
    \input{tab/r6_repeat_paraphrase}
\end{table}
\begin{table}[htbp]
    \centering
    \caption{\textbf{Repeated measurements with mechanics equations.} These rates do not identify how the model used the equations.}
    \label{tab:repeat-equations}
    \input{tab/r6_repeat_equations}
\end{table}
\begin{table}[htbp]
    \centering
    \caption{\textbf{Repeated measurements after a request to check remaining possibilities.} Repetition and total choice accuracy measure different behavior.}
    \label{tab:repeat-remaining}
    \input{tab/r6_repeat_remaining}
\end{table}
\begin{table}[htbp]
    \centering
    \caption{\textbf{Choices without an observed measurement.} Repetition is defined relative to the matched base test. This comparison does not isolate test labels from the rest of the image.}
    \label{tab:repeat-noimage}
    \input{tab/r6_repeat_noimage}
\end{table}
\FloatBarrier

\section{Resources and Materials}
\label{sec:resources}
The recorded study used up to four NVIDIA A100 PCIe GPUs with $40$\,GB each.
Smaller checkpoints used one device; Qwen 32B used two.
Inference used BF16 on these Ampere GPUs, with no training or quantization.
The recorded environment lists Python $3.12.3$, PyTorch $2.6.0$ with CUDA $12.4$, Transformers $4.51.3$, NumPy $2.5.2$, and Pillow $12.3.0$.
Pixtral uses eager attention because the tested vision implementation does not support the selected SDPA path.
The original V100 path was not exercised.

\begin{table}[htbp]
    \centering
    \caption{\textbf{Recorded request time, input length, and GPU memory.} Time and input tokens are averages per call. GPU GB is peak allocated memory on the monitored device. For Qwen 32B, this is not the total across its two GPUs. Its reasoning run also has a different mix of requests.}
    \label{tab:resources}
    \input{tab/r8_resources}
\end{table}
The direct runs sum to about $8.9$ hours of recorded request time and the reasoning runs to about $78.1$ hours.
These sums are not total GPU hours.
One model uses two devices, requests run concurrently, and loading and other overhead are excluded.
The execution notes report about $31.8$ hours from initial implementation to the last model job and approximately $141$\,GB of task storage, including checkpoints.
They also report $4{,}222$ pilot calls beyond the main study counts.
These are recorded project totals, not measurements of a deployable system's cost.
Pixtral's maximum direct input length reaches $4{,}245$ tokens, above the planned $4{,}096$ cap.
The decision was to preserve the original image resolution across models.
No model weights are included in the paper source package.



\subsection{Qualitative Example Selection}
\label{sec:qualitative-records}
The success in Table~\ref{tab:qualitative} uses the lower threshold from bounce family $015$.
Its two mass measurements leave possible rebound energies $\{1.82789,7.31157\}$ and $\{3.65576,14.6231\}$\,J.
The threshold is $2.74182$\,J.
One history therefore needs a bounce test and the other already implies \yes.
The unsuccessful example uses the lower threshold from sliding family $004$.
Figures~\ref{fig:teaser} and~\ref{fig:method} use the same family.
Its two friction measurements leave possible travel distances $\{0.801469,3.20589\}$ and $\{0.267156,1.06863\}$\,m.
The threshold is $0.534312$\,m.
For these two examples, action labels come from explicit final option letters.
The examples were chosen after evaluation to illustrate contrasting behaviors.

\subsection{Broader Impacts and Model Assistance}
\label{sec:ethics}
An agent that stops with insufficient evidence or repeats an uninformative test may waste resources or act on an unsupported physical prediction.
The present task can help measure such behavior under explicit assumptions.
Its schematic environments cannot establish safety in physical deployment.
The study includes no robot interaction, human participants, or personal data.
It trains no new model and releases no model weights.
Language models assisted with manuscript preparation, inspection of supplied files, and figure and LaTeX code.
The reported experimental values come from recorded model responses and stored metrics.
No new model evaluation was performed for this manuscript version.